\documentclass[10pt,twocolumn,letterpaper]{article}

\usepackage{cvpr}
\usepackage{times}
\usepackage{epsfig}
\usepackage{graphicx}
\usepackage{amsmath}
\usepackage{amssymb}
\usepackage{color}
\usepackage[pagebackref=true,breaklinks=true,letterpaper=true,colorlinks,bookmarks=false]{hyperref}
\usepackage[boxed]{algorithm2e}

\newcommand{\T}{\ensuremath{\mathcal{T}}}
\newcommand{\redbf}{\color{blue}\textbf}

\cvprfinalcopy
\def\cvprPaperID{2937} 

\ifcvprfinal\fi

\begin{document}
\title{Deformation Aware Image Compression}

\author{Tamar Rott Shaham and Tomer Michaeli\\
	Technion---Israel Institute of Technology, Haifa, Israel\\
	{\tt\small\{stamarot@campus,tomer.m@ee\}.technion.ac.il}
}

\maketitle
\thispagestyle{empty}

\begin{abstract}
Lossy compression algorithms aim to compactly encode images in a way which enables to restore them with minimal error. 
We show that a key limitation of existing algorithms is that they rely on error measures that are extremely sensitive to geometric deformations (\eg SSD, SSIM). These force the encoder to invest many bits in describing the exact geometry of every fine detail in the image, which is obviously wasteful, because the human visual system is 
indifferent to small local translations. Motivated by this observation, we propose a deformation-insensitive error measure that can be easily incorporated into any existing compression scheme. As we show, optimal compression under our criterion involves slightly deforming the input image such that it becomes more ``compressible''. Surprisingly, while these small deformations are barely noticeable, they enable the CODEC to preserve details that are otherwise completely lost. Our technique uses the CODEC as a ``black box'', thus allowing simple integration with arbitrary compression methods. Extensive experiments, including user studies, confirm that our approach 
significantly improves the visual quality of many CODECs. These include JPEG, JPEG~2000, WebP, BPG, and a recent deep-net method.

\end{abstract}

\section{Introduction}

\begin{figure}
	\centering
	\includegraphics[width=1\columnwidth]{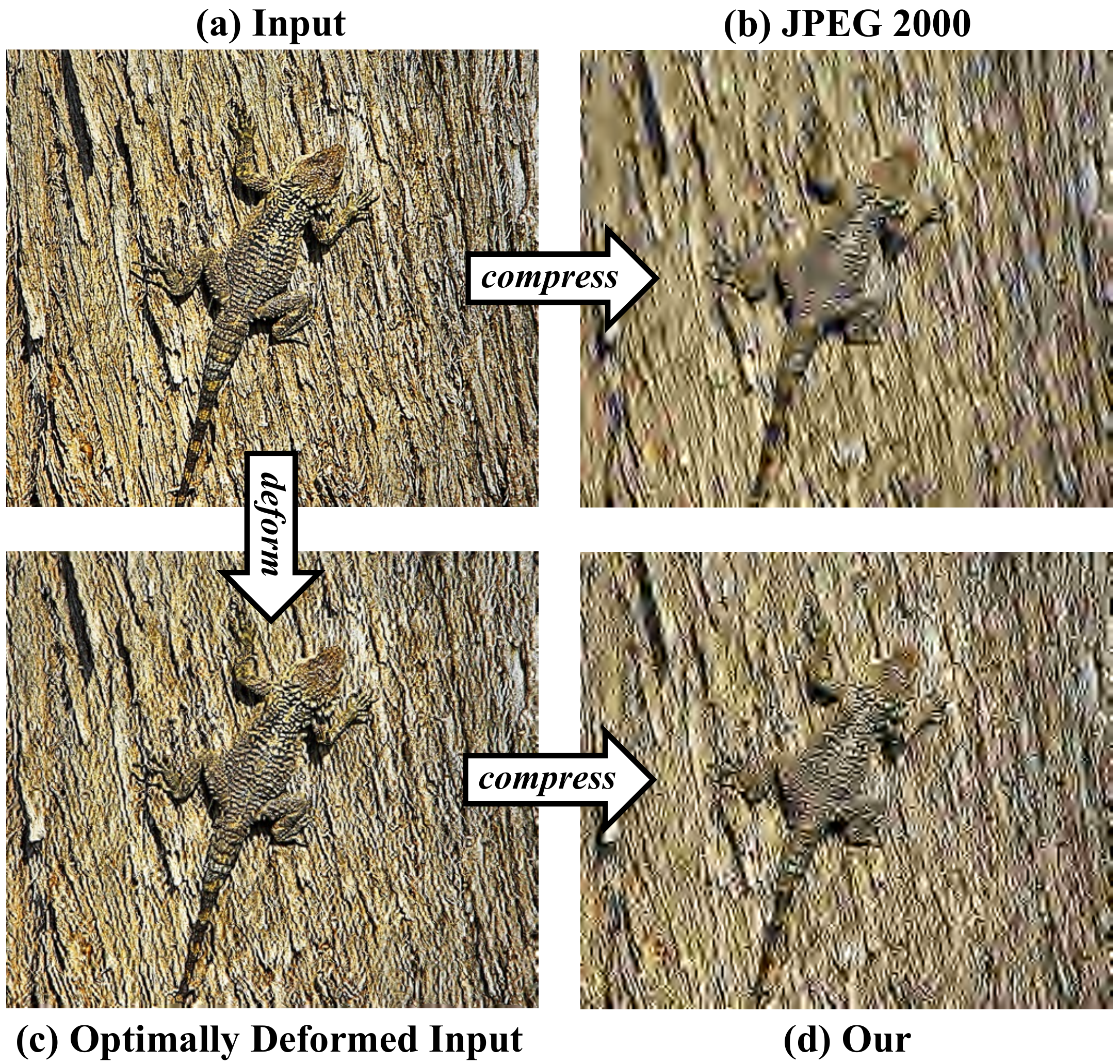}
	\caption{\textbf{Deformation aware image compression.} 
	Our algorithm seeks to minimize a deformation-insensitive error measure. This boils down to determining how to best deform the input image~(a) so as to make it more compressible~(c). By doing so, we trade a little geometric integrity with a significant gain in terms of preservation of visual information~(d) compared to regular compression~(b). Note that images~(b) and~(d) were obtained by compressing images~(a) and~(c), respectively, with the \emph{same} JPEG~2000 CODEC using the \emph{same} compression ratio of 150:1.}\label{Fig:Figure1}
\end{figure}

The last decades have seen an exponential rise in the popularity of mobile devices equipped with high-resolution cameras. To accommodate the numerous amounts of pictures captured by those devices on a daily basis, there is a crucial need for high quality compression algorithms. Indeed, while~20 megapixel images are becoming common (requiring~60 megabytes to store uncompressed), transmission and storage is often limited to less than 1~megabyte per image. At such high ratios, commonly used compression methods tend to discard important information from the image and produce visually unpleasing results.

In this paper, we propose a generic approach for boosting the visual quality of \emph{any} image compression method, by introducing deformations to the input image (see Fig.~\ref{Fig:Figure1}). Our algorithm uses the CODEC as a ``black box'' and is thus extremely simple to incorporate into arbitrary methods. Yet, it has a pronounced effect: At the same bit rate, we are able to achieve significantly better visual results.

Lossy compression schemes attempt to compactly encode images in a way which allows to restore them with minimal error. Over the years, most efforts to improve compression methods focused on seeking better image models. Examples include sparsity of image blocks in the DCT domain \cite{wallace1992jpeg}, recurrence of patches across different scales of the image~\cite{barnsley1990methods}, sparsity in the wavelet domain~\cite{skodras2001jpeg}, and smoothness (as exploited \eg by PDE based approaches~\cite{galic2005towards,schmaltz2014understanding}). However, the impact of new image models seems to be slowing down. Indeed, newer and more sophisticated priors now outperform their predecessors by small margins and only at high compression ratios \cite{Google2015WebP,Bellard2014BPG}.

Here, we take a different route. Rather than focusing on the image prior, we focus on the error criterion. Specifically, most compression methods seek to minimize some per-pixel distance (typically $\ell_2$) between the input image and the decoded image. Several attempts were also made to incorporate the 
SSIM index~\cite{wang2004image} as a fidelity criterion, leading to only modest improvement in visual quality \cite{richter2009ms,jiang2011jpeg}. We claim that the main limitation of most existing distance measures (including perceptual ones) is that they are very sensitive to slight misalignment of shapes and objects in the two images. Therefore, excelling under those criteria requires encoding the precise geometry of every fine detail in the image. But this is clearly wasteful, as the human visual system is not distracted by small geometric deformations, as long as the semantics of the scene is preserved.

Motivated by this insight, in this paper we propose a new error measure, which is insensitive to small smooth deformations. Our measure has two key advantages over other criteria: (i) it is very simple to incorporate into any compression method, and (ii) in the context of compression, it better correlates with human perception (as we confirm by user studies), and thus leads to a significant improvement in terms of detail preservation.

\begin{figure}[t]
	\centering
	\includegraphics[width=0.98\columnwidth, trim=0cm 0cm 0cm 0cm, clip]{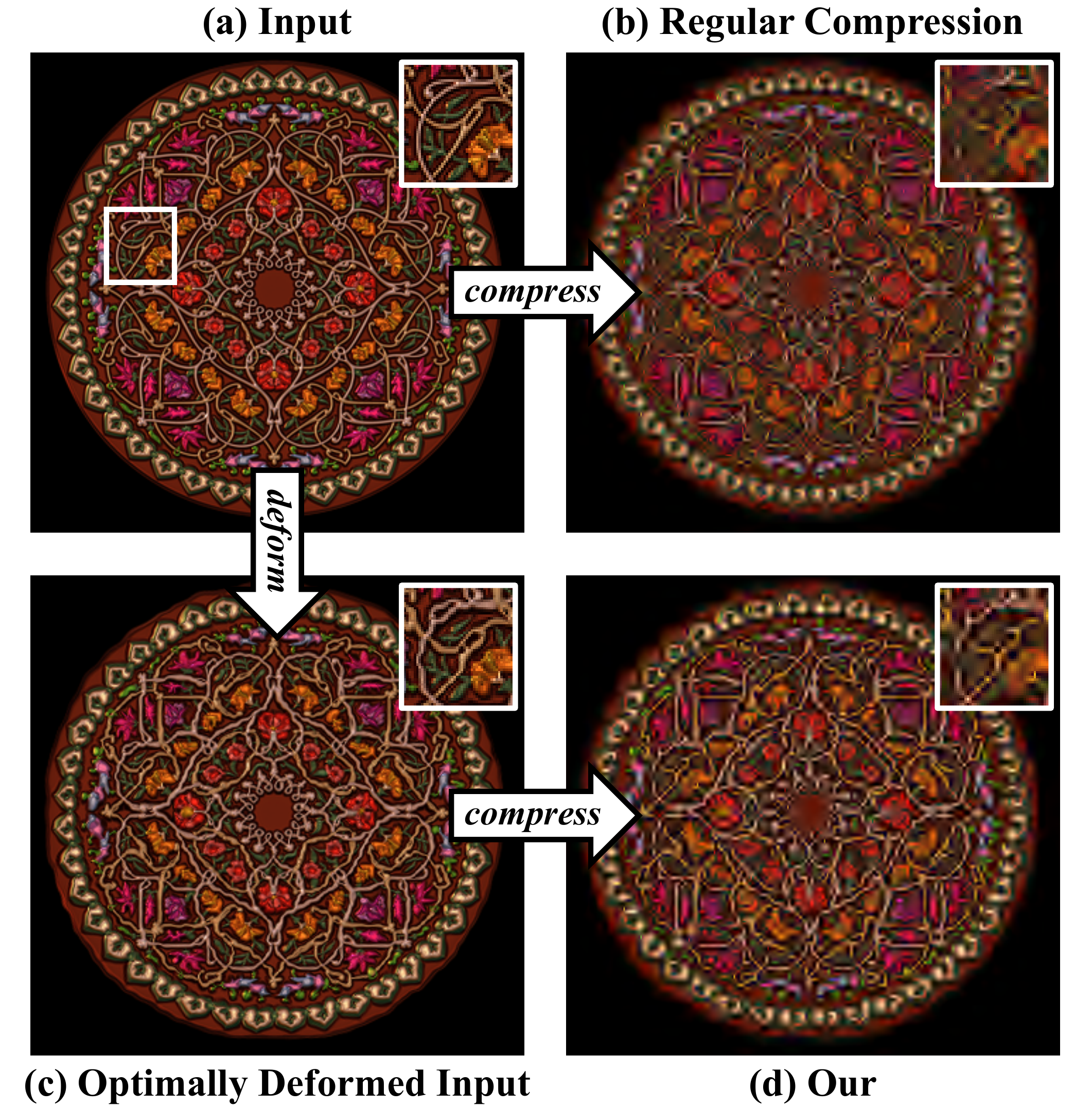}
	\caption{\textbf{The effect of deformation.} When compressing the input image~(a) using the Global Thresholding method~\cite{misiti2013wavelets} at a ratio of 40:1, delicate structures are completely lost (b). Interestingly, by making those structures a bit wiggly~(c) they can be significantly better preserved during compression~(d).}\label{Fig:DragonFly}
\end{figure}

As we show, optimal compression under our criterion boils down to determining how to best deform the input image such that it becomes more ``compressible''. 
This is illustrated in Figs.~\ref{Fig:Figure1} and~\ref{Fig:DragonFly} for the JPEG~2000~\cite{skodras2001jpeg} and the Global Thresholding~\cite{strang1996wavelets} compression methods. As can be seen, by introducing very minor deformations, we are able to make the compression scheme preserve delicate features that are otherwise completely lost. Note that this effect is achieved without increasing the bit rate. 
In other words, rather than discarding textures and small objects to meet the bit budget, we geometrically modify them such that they can be better encoded with the same number of bits.

\begin{figure}[t]
	\centering
	\includegraphics[width=1\columnwidth, trim=0cm 0cm 0cm 0cm, clip]{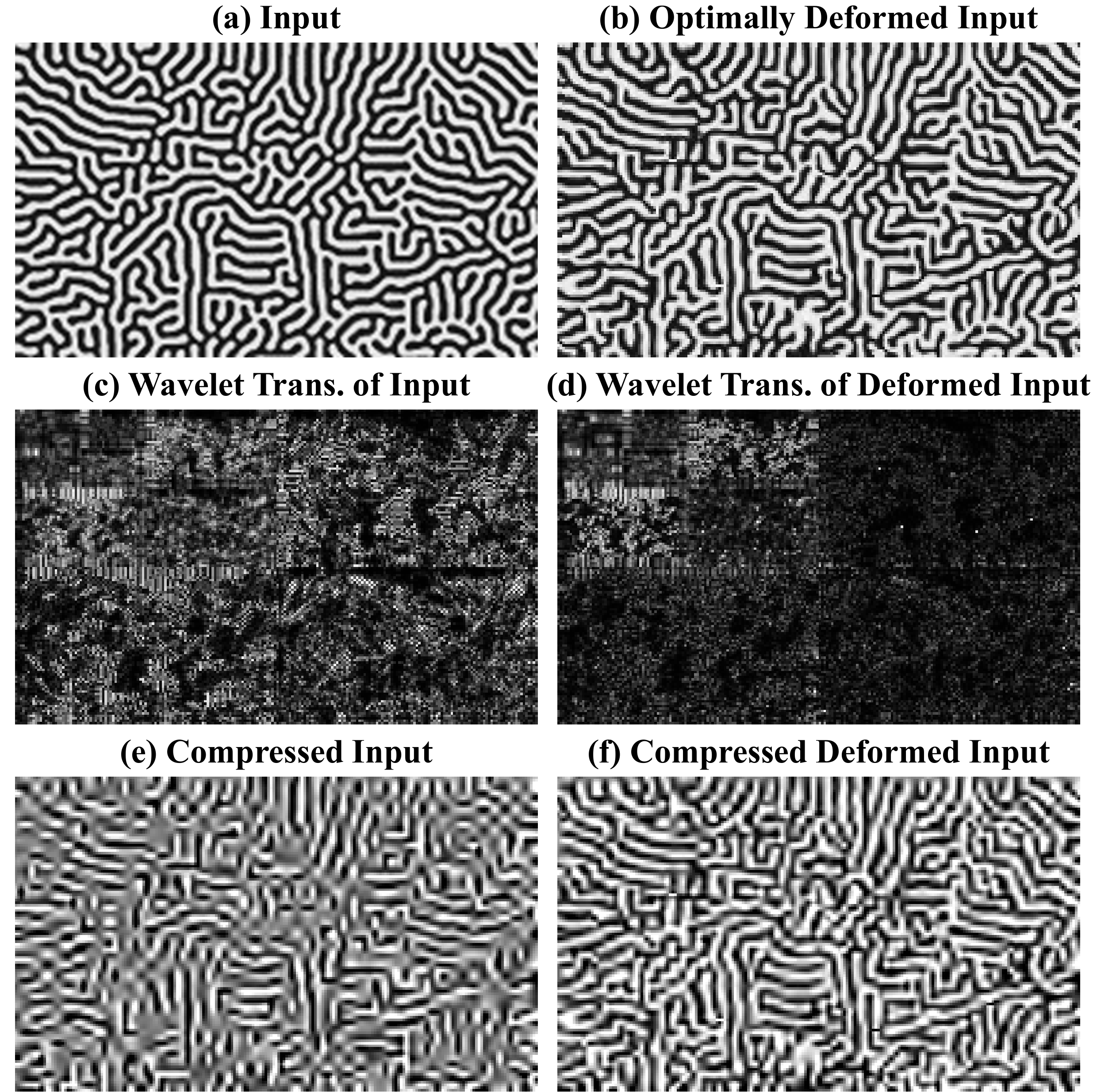}
	\caption{\textbf{Deformation aware compression via Subband Thresholding.} The input image~(a) contains many strong curved edges, so that its wavelet transform (c) is not very sparse. This causes Subband Thresholding compression~\cite{strang1996wavelets} at a ratio of 25:1, to produce a very blurry result~(e). However, by introducing a minor geometric deformation~(b), we are able to make the wavelet transform of the image much sparser~(d). This allows the compression algorithm to preserve most of the structures in the image, at the same bit budget~(f). This principle applies to arbitrary image priors, and not only those based on wavelet sparsity.}
	\label{Fig:Pattern}
\end{figure}

The surprising success of our approach can be attributed to an interesting phenomenon recently observed in \cite{RottShaham2016visualizing}. That is, by introducing small geometric deformations, it is usually possible to significantly increase the likelihood of any natural image under any given prior. The implication of this effect on compression is striking. For example, compression algorithms that exploit sparsity in the wavelet domain (\eg JPEG~2000), discard the small wavelet coefficients of the image. At high compression ratios, this causes fine details to fade, as demonstrated in Fig.~\ref{Fig:Pattern} for the Subband Thresholding compression method \cite{strang1996wavelets}. However, as can be seen in Fig.~\ref{Fig:Pattern}(b),(d), it takes only a small deformation to make the wavelet transform of the image significantly sparser. Thus, by slightly sacrificing geometric integrity, we substantially improve the ability of the compression algorithm to preserve details, as seen in Fig.~\ref{Fig:Pattern}(f).

\section{Related Work}


The most popular error measure in image compression is the squared $\ell_2$ distance. This criterion is mathematically convenient, being convex and differentiable, but is unarguably not well correlated with subjective human perception of image quality \cite{wang2009mean,lin2011perceptual,zhang2012comprehensive}.

There exist many fidelity criteria that are better correlated with human perception. A few examples are SSIM~\cite{wang2004image}, MS-SSIM~\cite{wang2003multiscale}, CW-SSIM~\cite{sampat2009complex}, IFC~\cite{sheikh2005information}, VIF~\cite{sheikh2006image}, and FSIM~\cite{zhang2011fsim}.
Recently, several perceptual loss functions have been proposed, which measures the 
similarity between deep feature maps 
(mostly of the VGG net). These measures were shown to lead to pleasing visual results in a variety of low-level vision tasks, including super-resolution \cite{ledig2016photo,johnson2016perceptual} and style transfer \cite{luan2017deep}. Perceptual losses were also incorporated with generative adversarial networks (GANs), allowing to achieve high-quality super-resolution~\cite{ledig2016photo} and compression \cite{waveone2017} results.

While these criteria better match human perception,
their  majority lack a crucial property for perceptual compression: \emph{deformation invariance}. Namely, they do not tolerate small misalignment of objects, and thus necessitate the encoder to invest many bits in encoding the fine geometry of every feature in the image. Indeed, several attempts to incorporate the SSIM criterion into JPEG and JPEG~2000~\cite{richter2009ms,jiang2011jpeg} and into video coding~\cite{wang2010ssim,ou2011ssim,wang2012ssim}, led to rather modest improvements in visual quality.

Another drawback of existing error measures is that they
are difficult to incorporate into arbitrary compression schemes. Namely, as opposed to the $\ell_2$ distance, rate-distortion optimization under those measures cannot be done analytically and thus requires various approximations. Thus, even when solutions exist \cite{richter2009ms,jiang2011jpeg,ou2011ssim}, they are quite specific and cannot be easily extended to other compression standards. Furthermore, from an end-user viewpoint, those solutions require a specialized CODEC. Our solution, on the other hand, can work with any existing CODEC. Specifically, we use the CODEC as a ``black box'', to generate a preprocessed (deformed) image. This image can then be compressed and decompressed using the original CODEC, without any modification. 

Our method is related to a recent line of work on using deformations for idealizing images \cite{dekel2015revealing,wadhwa2015deviation,RottShaham2016visualizing}. These papers introduced the idea of deforming images as a means for making them better comply with some prior model. Here, we harness this idea for improving image compression. Namely, we propose an error criterion which measures similarity up to small deformations. Thus, compression under our criterion, reduces to determining how the input image should be deformed so that it is more compressible.


Note that the idea of measuring image similarity up to deformation has been proposed in the context of image recognition~\cite{moghaddam1996bayesian} and face recognition~\cite{wagner2012toward}
. However, this approach has never been exploited for image compression.

\section{Deformation Aware Compression}

Modern compression schemes involve a procedure known as rate-distortion optimization. Namely, during compression, the algorithm adaptively selects where to invest more bits so as to minimize the distortion between the input image~$y$ and its compressed version~$x$, while conforming to a total bit rate constraint of $\varepsilon$ bits per pixel. This can be formulated as the optimization problem
\begin{equation}\label{eq:Rate Distortion}
\min_x  d(x,y) \quad \text{s.t.}\quad R(x) \leq \varepsilon,
\end{equation}
where $d(\cdot , \cdot)$ is some distortion measure that quantifies the dissimilarity between $x$ and $y$, and $R(x)$ is the rate required to encode $x$.

The most popular distortion measure is the sum of squared differences (SSD), \ie the square $\ell_2$ error norm
\begin{equation}\label{eq:SSD}
d_{\text{SSD}}(x,y) = \| x-y \|^2.
\end{equation}
The SSD is a per-pixel criterion, and is therefore extremely sensitive to slight misalignment or deformation of objects. For that reason, when the bit budget $\varepsilon$ is low (\ie high compression ratio), the compression process completely removes or blurs out certain structures in the image.

\begin{figure*}[t]
	\centering
	\includegraphics[width=0.95\textwidth, trim=0cm 0cm 0cm 0cm, clip]{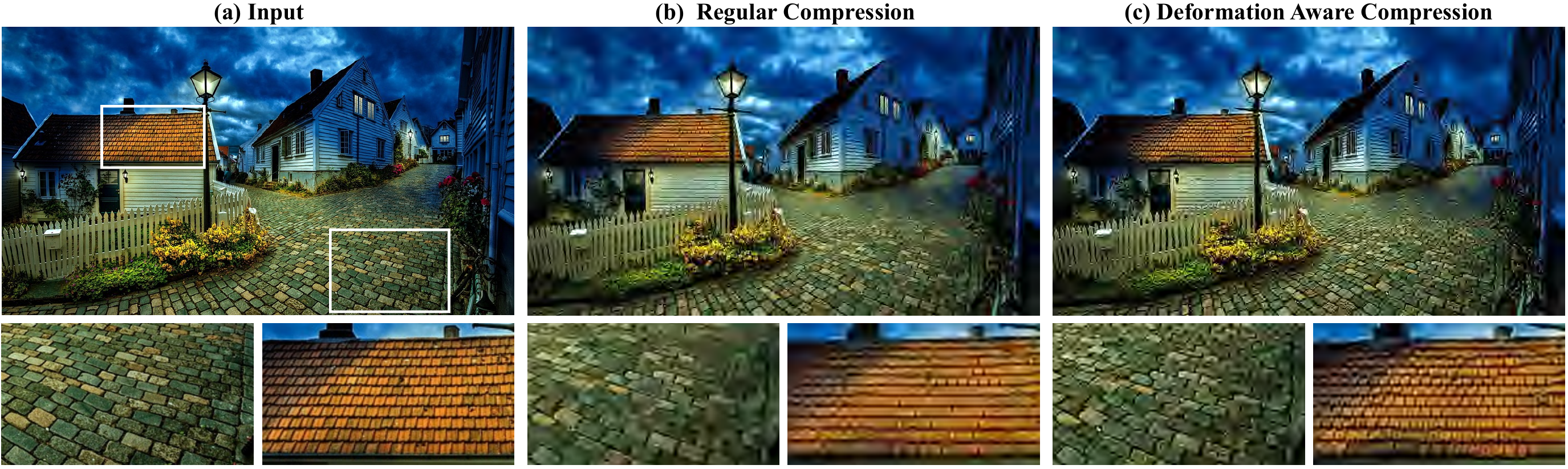}
	\begin{tabular}[width=1\textwidth]{c|c|c|c|c|c|c|c|c|c|c|}
		\cline{2-11}
		& \multicolumn{6}{ c| }{Similarity scores (higher $=$ similar)} &
		\multicolumn{4}{ c| }{Dissimilarity scores (lower $=$ similar)} \\ 	\cline{2-11}
		&SSIM & MS-SSIM & CW-SSIM & IFC & VIF & FSIM & VGG$_{2,2}$ &  VGG$_{5,4}$ & SSD & DASSD \textbf{(our)} \\ \cline{1-11}
		
		\multicolumn{1}{ |c| }{JPEG~2000} & \redbf{0.696} &\redbf{ 0.837} & \redbf{0.999} & \redbf{1.216} & \redbf{0.124} & \redbf{0.857} & \redbf{0.173} & \redbf{2.023} & \redbf{545} & 328 \\ \cline{1-11}
		\multicolumn{1}{ |c| }{Our} 	  & 0.685         & 0.810	      & 0.996         & 1.014	      & 0.104         & 0.848	      & 0.252         & 2.347         &  639        & \redbf{284} \\ \cline{1-9}
		\hline	
	\end{tabular}
	\caption{\textbf{Comparison between error criteria.} We measured the similarity between the input image~(a) and the compressed images (b) and (c), according to various fidelity criteria. We indicate in bold which compression result has been ranked more similar: our deformation aware compression~(c), or the regular JPEG~2000 compression~(b) (both compressed at a ratio of 200:1). As can be seen, all error criteria other than DASSD rank our result as less similar to the input image. This is despite the fact that it preserves much more visual information.} \label{Fig:DAMSE}
\end{figure*} 	

As human observers are indifferent to slight local translations, 
here we propose a deformation insensitive version of the SSD measure. We consider two images $x$ and $y$ to be similar, if there exists a smooth deformation $\T$ such that $x$ and $\T\{y\}$ are similar. More concretely, we define the deformation aware SSD (DASSD) between $x$ and $y$ as
\begin{equation}\label{eq:Deformation allowing Distortion}
d_{\text{DASSD}}(x,y) = \min_{\T} \| x-\T \{y\} \|^2 +\lambda\,\psi(\T),
\end{equation}
where the term $\psi(\T)$ penalizes for non-smooth deformations. Note that computing the DASSD requires solving an optical-flow problem \cite{horn1981determining} to determine how to best warp~$y$ onto~$x$. 
Once the optimal deformation is determined, the DASSD is defined as the SSD between~$x$ and the warped~$y$, plus a term that quantifies the roughness of the flow field. The parameter $\lambda$ controls the tradeoff between the two terms. Therefore, the DASSD is large if the best warped~$y$ is not similar to~$x$, or if the deformation required to make~$y$ similar to~$x$ is not smooth (or both).

To allow for complex deformations, we use a nonparametric flow field $(u,v)$, namely
\begin{equation}\label{eq:deformation}
\T\{ y \} ( \xi , \eta ) = y( \xi +u(\xi , \eta ), \eta +v( \xi , \eta )).
\end{equation}
We define the penalty $\psi(\T)$ to be a weighted Horn and Schunk regularizer \cite{horn1981determining},
\begin{equation}\label{eq:rho}
\psi(\T) = \iint w(\xi,\eta) \left(\|\nabla u(\xi,\eta)\| ^2 +\| \nabla v( \xi , \eta ) \| ^2 \right) d\xi d\eta,
\end{equation}
where $\nabla = (\tfrac{\partial}{\partial\xi},\tfrac{\partial}{\partial\eta})$ and $w(\xi,\eta)$ is a weight map that puts higher penalty on salient regions (see Sec.~\ref{RegM}).

Figure~\ref{Fig:DAMSE} illustrates the advantage of our DASSD criterion over other similarity criteria. In this example, our method clearly preserves more details than the original JPEG~2000 compression. 
This is captured by our DASSD measure, which ranks our compressed image as more similar to the input image than the regular JPEG~2000 result (here we used a constant regularization map $w(\xi,\eta)=1$). However, the other similarity measures are very sensitive to misalignment of objects, and thus all of them rank our result as less similar to the input image. 
Please see a comparison of all similarity measures on all the images in this paper in the Supplementary Material. 

\begin{figure}[t]
	\centering
	\includegraphics[width=1\columnwidth, trim=0cm 0cm 0cm 0cm, clip]{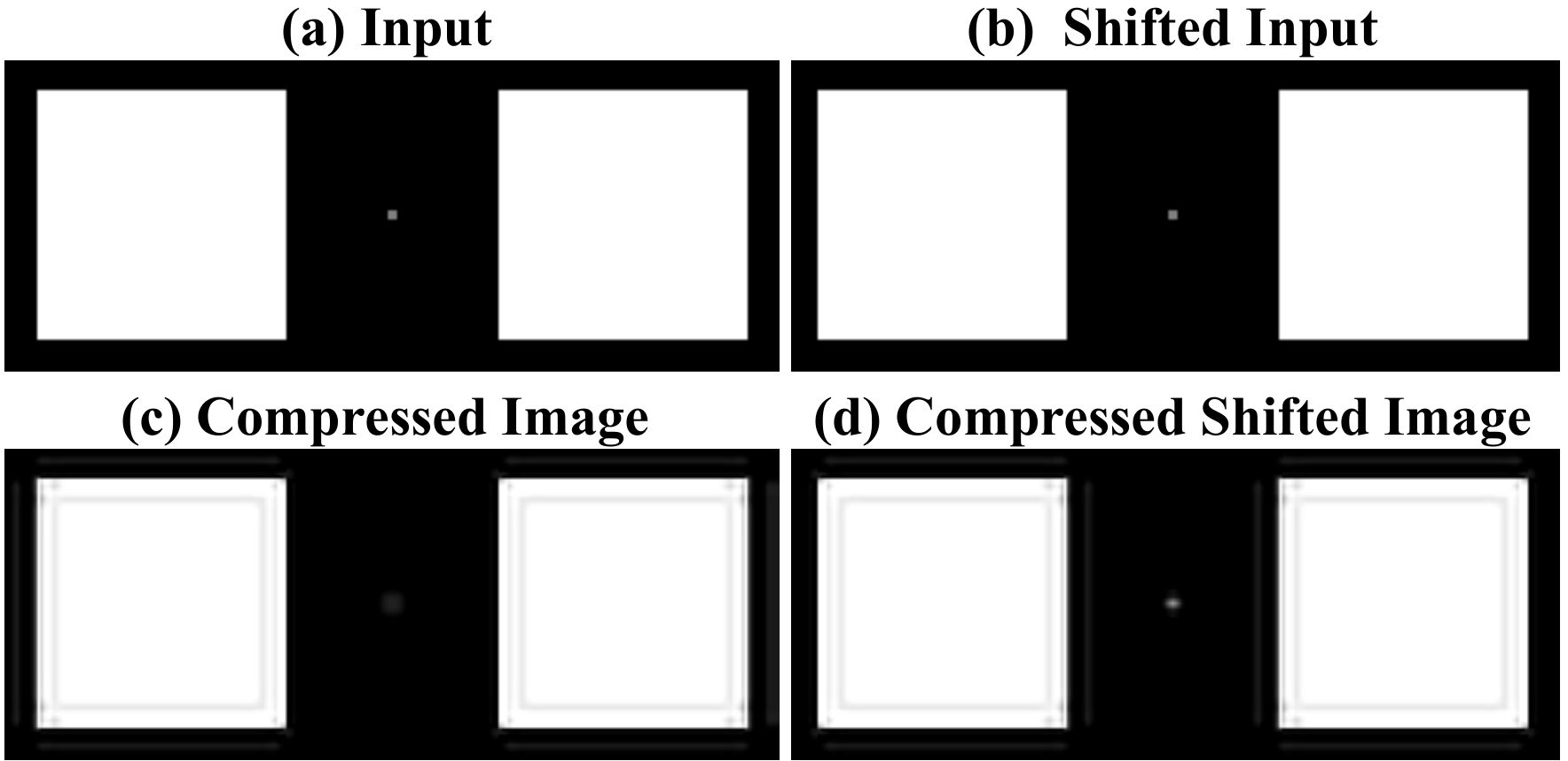}
	\caption{\textbf{The effect of global translation.} Compressing the image~(a) using JPEG~2000 at a ratio of 75:1, causes the small square in the middle to disappear~(c). However, by shifting the image only two pixels to the left~(b), compression at the same ratio keeps the small square intact~(d).}\label{Fig:Synthetic}
\end{figure}

Our measure is obviously the least sensitive to smooth deformations. But why should deformation invariance improve compression? Lossy image compression schemes are usually not translation invariant. That is, compressing a shifted version of an image, gives an entirely different result than shifting the compressed image. This is demonstrated in Fig.~\ref{Fig:Synthetic} for the JPEG~2000 standard. While the input image and its shifted version look perfectly identical to a human observer, their compressed versions look very different. In one of them the small square in the middle is preserved, and in the other it is not. As opposed to SSD, our deformation aware criterion prefers the result in which the small square is preserved: The DASSD between~(a) and~(d) is 3\% lower than the DASSD between (a) and (c), while the SSD between (a) and (d) is 16 times larger than the SSD between (a) and (c). This intuition can be extended to local translations. For example, to preserve the pattern of the Mandala in Fig.~\ref{Fig:DragonFly}, it is necessary to make lines a bit wiggly. These small translations make the image more compressible, thus leading to better visual quality at the same bit rate.

\section{Algorithm}

\begin{figure}[t]
	\centering
	\includegraphics[width=0.98\columnwidth, trim=0cm 0cm 0cm 0cm, clip]{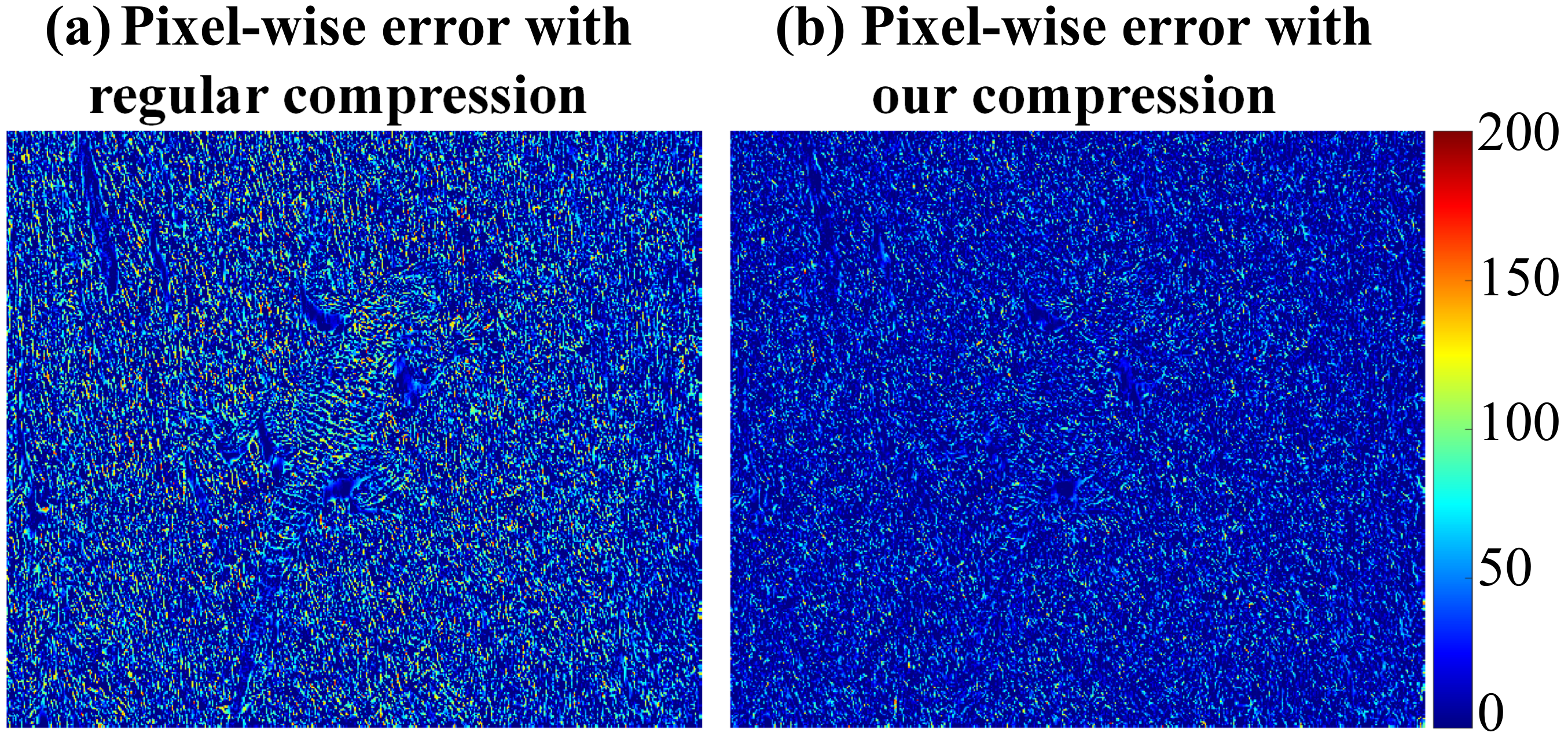}
	\caption{{\textbf{Pixel-wise compression error.} Our algorithm deforms images so as to make them more “compressible” in the $\ell_2$ sense. The effect is that the error obtained in compressing the deformed image~(b) is significantly lower than the error obtained in compressing the original image~(a). These results correspond to the Lizard images of Fig.~\ref{Fig:Figure1}(c) and Fig.~\ref{Fig:Figure1}(a), respectively.}} \label{PixelWiseMSE}
\end{figure}

Substituting $d_{\text{DASSD}}$ of~\eqref{eq:Deformation allowing Distortion} into~\eqref{eq:Rate Distortion}, we obtain the optimization problem
\begin{equation}\label{eq:deformation aware Rate-Distortion}
\min_{x,\T} \| \T \{y\}-x \|^2 +\lambda\,\psi(\T) \quad \text{s.t.} \quad R(x) \leq \varepsilon.
\end{equation}
That is, we need to simultaneously determine a compressed image~$x$ (represented with no more than~$\varepsilon$ bits per pixel) and a geometric deformation~$\T$, such that~$x$ is similar to the deformed image $\T\{y\}$ rather than to~$y$ itself. In other words, we seek how to deform the input image~$y$, such that $\T\{y\}$ can be compressed with smaller SSD error under the same bit budget. This is illustrated in Fig.~\ref{PixelWiseMSE}. 

To solve problem \eqref{eq:deformation aware Rate-Distortion} we alternate between minimizing the objective w.r.t.\@ $x$ while holding $\T$ fixed and vice versa.

\vspace{0.2cm}
\noindent$\boldsymbol{x}$\textbf{-step:}
When $\T$ is fixed, $\psi(\T)$ can be discarded, so that~\eqref{eq:deformation aware Rate-Distortion} simplifies to
\begin{equation}\label{eq:x-stem}
   \min_x \| x - \T \{y\} \|^2 \quad \text{s.t.} \quad R(x)\leq\varepsilon.
\end{equation}
This is a standard rate distortion 
problem, but for compressing the deformed image $\T\{y\}$ rather than the input image~$y$. 

\vspace{0.2cm}
\noindent{$\boldsymbol{\T}$\textbf{-step:}} When $x$ is fixed, the bit rate $R(x)$ is constant, and~\eqref{eq:deformation aware Rate-Distortion} reduces to
\begin{equation}\label{eq:T-step}
   \min_{\T} \| x - \T \{y\} \|^2 +\lambda\,\psi(\T).
\end{equation}
This is an optical flow problem \cite{horn1981determining} for determining how to best warp the input image $y$ onto the compressed image $x$. Here we use the iteratively re-weighted least-squares (IRLS) algorithm proposed in~\cite{liu2009beyond}.

Thus, as summarized in Alg.~\ref{alg:DeformationAwareCompression}, our algorithm iterates between two simple steps: Compressing the current deformed input image~$\T\{y\}$ to obtain~$x$, 
and computing the optical flow between $x$ and $y$ to update~$\T$. Note that the $x$-step can be done with any CODEC. This allows integrating our algorithm with arbitrary compression methods.

\begin{figure}[t]
	\centering
	\includegraphics[width=1\columnwidth, trim=0cm 0cm 0cm 0cm, clip]{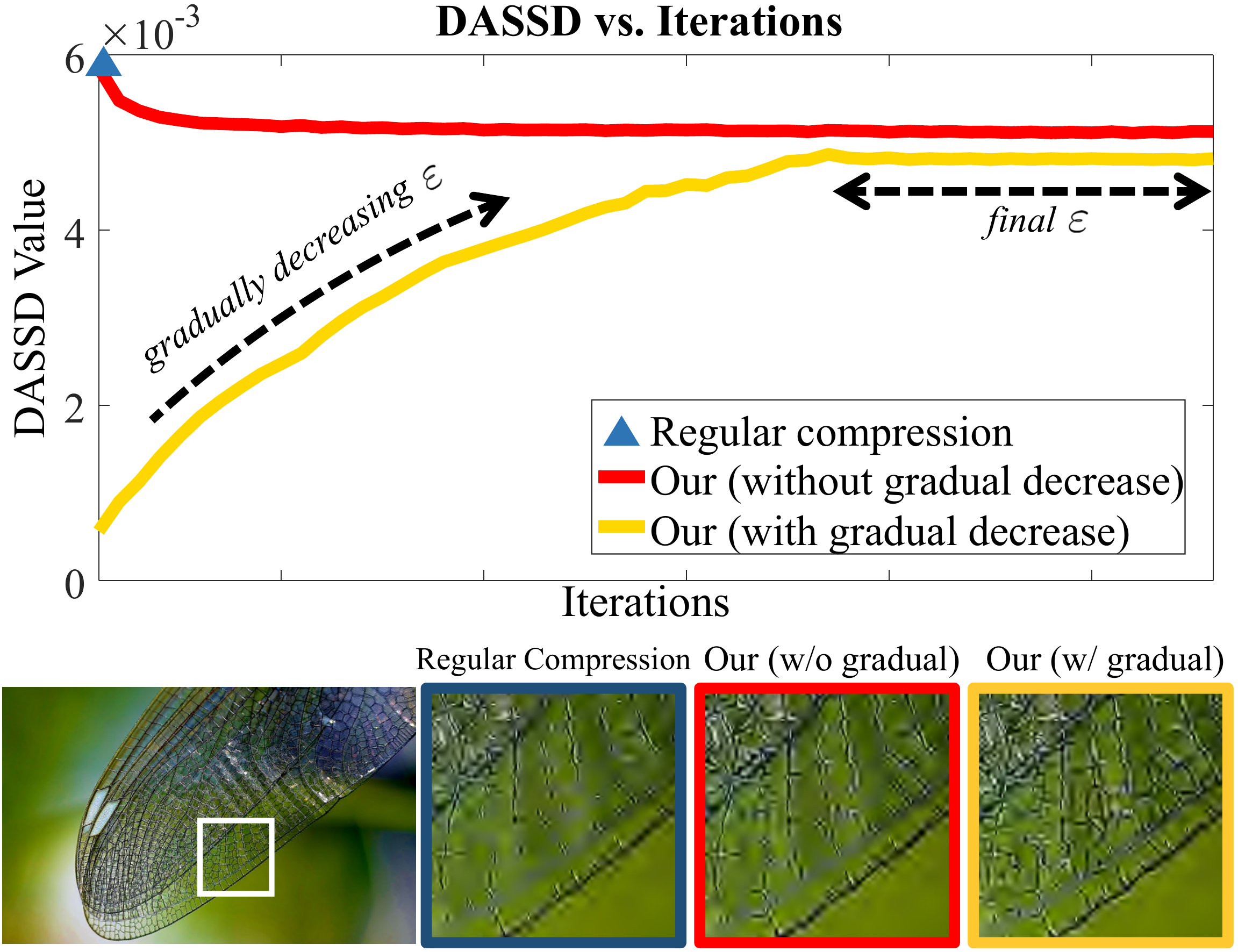}
	\caption{\textbf{The effect of gradual bit rate decrease.} Our algorithm iterates between compression and optical flow estimation. When the compression step is performed with the final desired bit rate (red box), the algorithm fails to restore all the fine structures on the Dragonfly's wing, that are lost with regular compression (blue box). However, when starting with a large bit rate and progressively decreasing it, our algorithm manages to preserve most of the wing's delicate textures (yellow box). This is correlated with the final DASSD value, which is 10\% lower when using the gradual scheme (yellow line), compared to the direct scheme (red line). We used the JPEG~2000 standard with a compression ratio of 150:1.}\label{Fig:GradualEffect}
\end{figure}

\begin{algorithm}[t]
	\KwIn{Image $y$, bit budget $\varepsilon$}
	\KwOut{Compressed image $x$}
	Initialize $\T$ to the identity mapping and $\tilde{\varepsilon}$ to be large 
	\\
	Compute the local regularization weight map $w$
	\\
	\While{$\tilde{\varepsilon}>\varepsilon$}{
		$x \, \leftarrow \texttt{Compress}(\T\{y\},\tilde{\varepsilon})$ \hspace{0.07cm} {\footnotesize{/* bit budget of $\tilde{\varepsilon}$ */}}
		\\
		$\T \leftarrow \texttt{OpticalFlow}(y, x)$ \hspace{0.07cm}{\footnotesize{/* with weight map $w$ */}}
		\\
		\texttt{decrease} $\tilde{\varepsilon}$
	}
	\caption{Deformation Aware Compression.}\label{alg:DeformationAwareCompression}
\end{algorithm}

\begin{figure*}[t]
	\centering
	\includegraphics[width=1\textwidth, trim=0cm 0cm 0cm 0cm, clip]{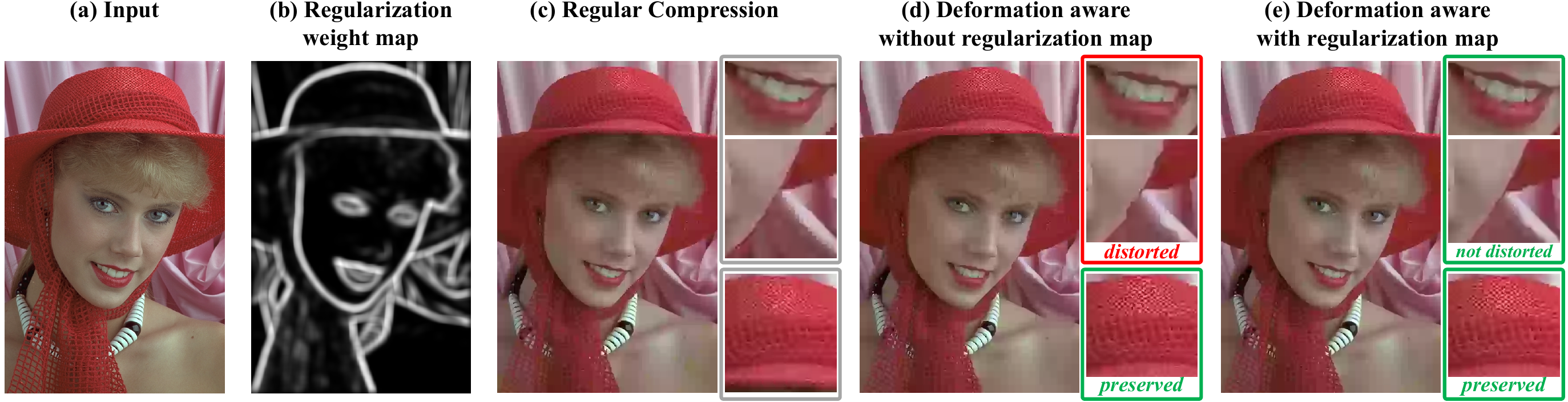}
	\caption{{\textbf{The effect of the regularization weight map.} Applying our scheme (with WebP compression at a ratio of 250:1) on image (a) without adaptive regularization (constant $w(\xi,\eta)$), results in image (d) in which the the hat's texture is better preserved, compared to the regular WebP compression (c), but the woman's lips are distorted unrealistically. Using the local regularization weight map (b), results in image (e) in which the boundaries of the lips are not distorted, yet the texture on the hat is still sharper than with the regular compression~(c).}} \label{Fig:AlphaMap}
\end{figure*}

To prevent the algorithm from getting trapped in a bad local minimum, we start with a large bit budget $\varepsilon$, and gradually decrease it along the iterations until we reach the desired budget. This helps to avoid the following situation. Suppose that at some stage, the compression ($x$-step) removes some structure from the image, so that it appears in~$\T\{y\}$ but not in~$x$. In that case, the optical flow ($\T$-step) cannot determine how to best deform this structure so as to encourage the compression to preserve it in the next iteration. Our gradual process overcomes this issue by allowing the deformation to gradually adapt to the small structures before they disappear. Therefore, as we show in Fig.~\ref{Fig:GradualEffect}, this process leads to better detail preservation. Indeed, with the gradual scheme, the algorithm converges to a DASSD value which is 10\% lower than without the gradual scheme. Note that the low DASSD values at early iterations are a result of using a high bit rate. As the bit rate decreases, the DASSD values increase (but reach a lower value at the final bit rate, than without the gradual process).

\vspace{0.2cm}
\noindent\textbf{Constructing the regularization weight map} \label{RegM}
To ensure good visual quality, we need to prevent extreme distortions at regions which capture the observer's attention. In particular, humans are very sensitive to the outline of objects. We therefore construct the weight map $w(\xi,\eta)$ in \eqref{eq:rho} as
\begin{equation} \label{eq:RegularizationMap}
	w(\xi,\eta) = 1+\alpha \cdot \, G(\xi,\eta) * E(\xi,\eta),
\end{equation}
where $E$ is an edge map obtained by applying the edge detector of \cite{dollar2013structured} on the input image $y$, $G$ is a Gaussian filter with 
$\sigma = 10$, `$*$' denotes convolution, and $\alpha$ is a parameter that controls the strength of the varying regularization. 
As can be seen from Fig.~\ref{Fig:AlphaMap}, using local regularization is essential for avoiding distracting artifacts. With a global regularization (constant $w$), object boundaries are distorted unrealistically (\eg the woman's lips). By introducing a spatially varying regularization, lines and boundaries are not distorted, yet textures (\eg the hat) are still allowed to deform and are thus better preserved during compression.

\section{Experiments}


\begin{figure*}[]
	\centering
	\includegraphics[width=0.95\textwidth, trim=0cm 0cm 0cm 0cm, clip]{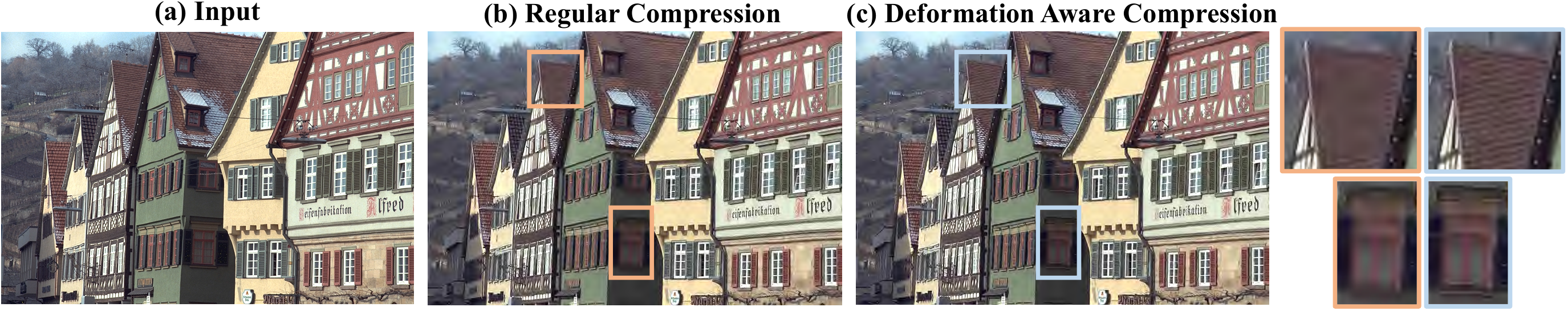}
	\caption{\textbf{JPEG~2000 \cite{skodras2001jpeg}.} Compression of Houses at a ratio of 50:1.
	}\label{Fig:JPEG2000}
	\medskip
	\centering
	\includegraphics[width=0.95\textwidth, trim=0cm 0cm 0cm 0cm, clip]{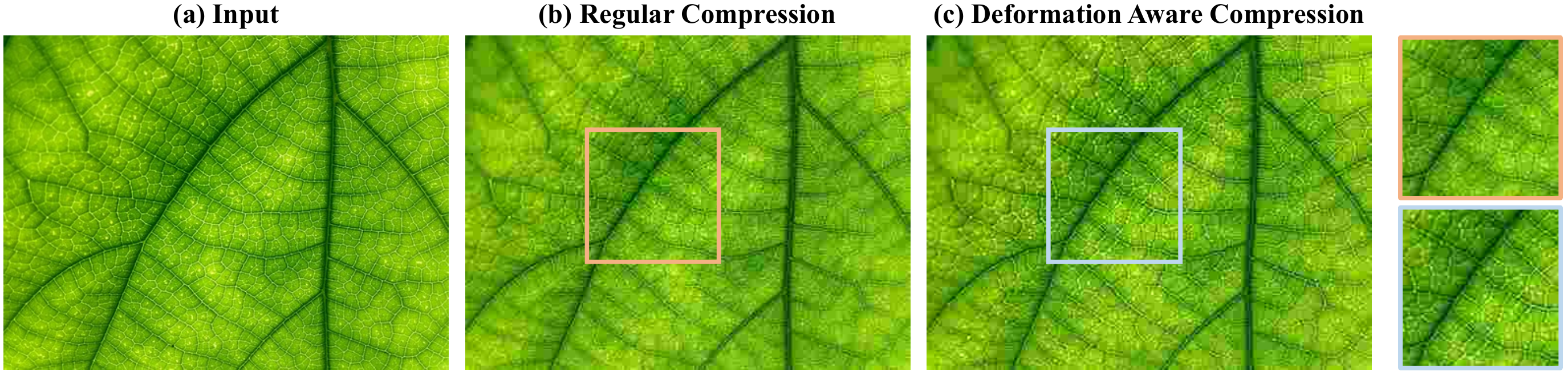}
	\caption{\textbf{JPEG \cite{wallace1992jpeg}.} Compression of Leaf at a ratio of 50:1.
	}\label{Fig:JPEG}
	\medskip
	\centering
	\includegraphics[width=0.95\textwidth, trim=0cm 0cm 0cm 0cm, clip]{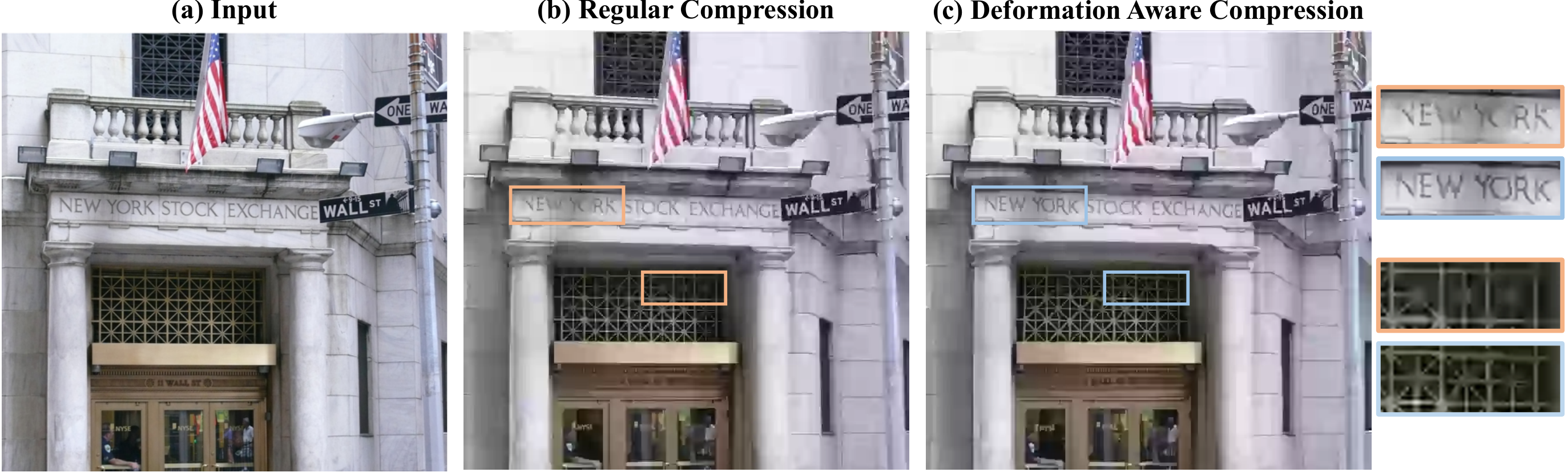}
	\caption{\textbf{WebP \cite{Google2015WebP}.} Compression of the New York Stock Exchange building at a ratio of 110:1.}\label{Fig:WebP}
	\medskip
	\centering
	\includegraphics[width=0.95\textwidth, trim=0cm 0cm 0cm 0cm, clip]{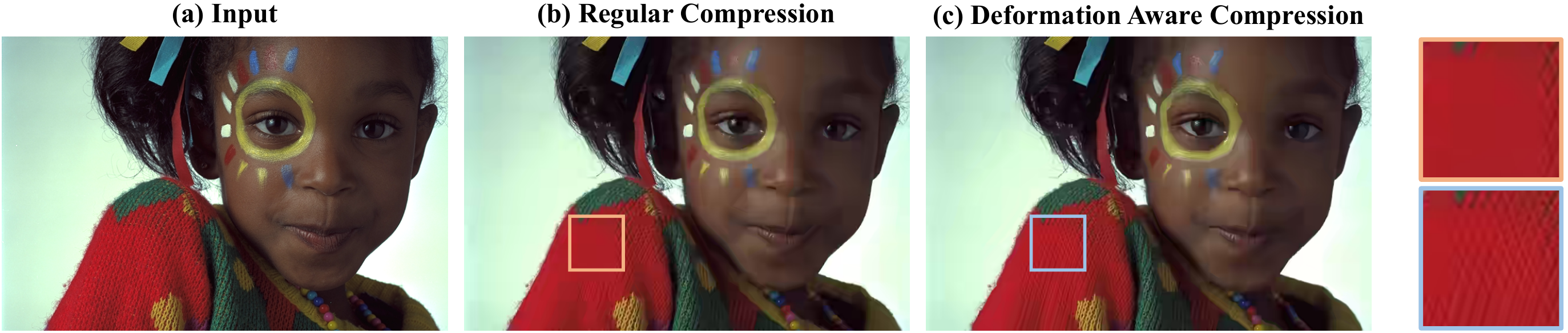}
	\caption{\textbf{BPG \cite{Bellard2014BPG}.} Compression of Girl at a ratio of 220:1.}\label{Fig:BPG}
	\medskip
	\centering
	\includegraphics[width=0.95\textwidth, trim=0cm 0cm 0cm 0cm, clip]{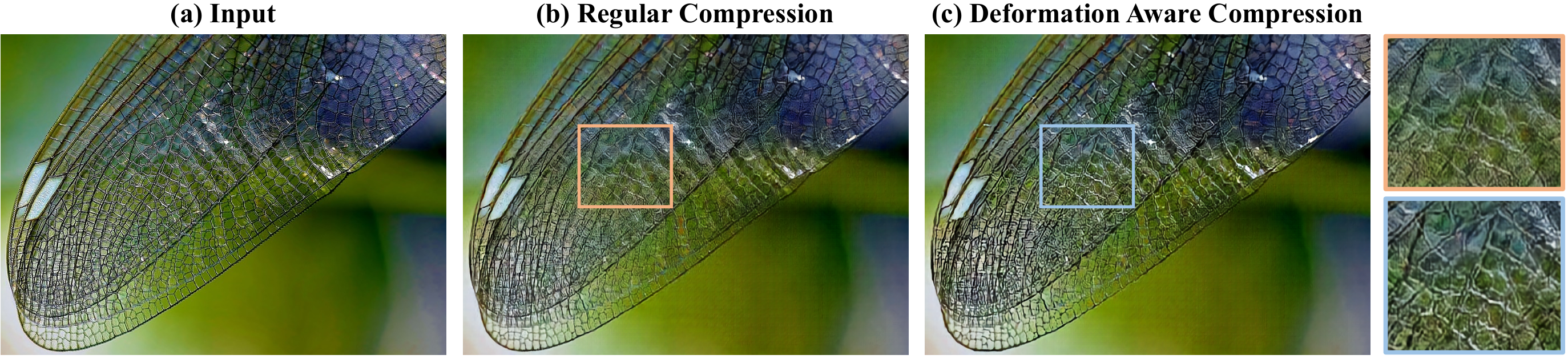}
	\caption{\textbf{Deep Coding \cite{toderici2016full}.} Compression of Dragonfly's wings at a ratio of 48:1.}\label{Fig:NN}
\end{figure*}

We tested our approach with JPEG~\cite{wallace1992jpeg}, JPEG~2000~\cite{skodras2001jpeg}, WebP~\cite{Google2015WebP}, BPG~\cite{Bellard2014BPG}, the deep-net based CODEC of~\cite{toderici2016full}, Subband Thresholding \cite{strang1996wavelets} and Global Thresholding \cite{misiti2013wavelets}, on images from the Berkeley segmentation dataset \cite{MartinFTM01}, the Kodak dataset~\cite{Kodak} and the Web (please see many more results in the Supplementary Material). For the gradual process, we kept the compression ratio fixed for the first 10 iterations, then increased it every 5 iterations for the next 25 iterations, and then increased it every single iteration until reaching the desired rate. For JPEG~2000, we started at a compression ratio of 20:1 and increased it by steps of 5. For Subband Thresholding and Global Thresholding we started at a ratio of 5:1 and increased it by steps of 1. For the deep coding algorithm, we started with a bit rate of 0.75 bits per pixel (BPP), and then decreased it by steps of 0.125 BPP.
In JPEG, BPG and WebP, the user specifies a quality parameter rather than the desired compression rate. Thus, for JPEG and WebP, we started at a quality of 50 and decreased it by steps of 1. For BPG we started at a quality index of 30 (here a lower index corresponds to better quality) and increased it by steps of~1. In each optical flow step, we used the flow from the previous iteration as initialization. We found this leads to better convergence. All warped images were produced with bicubic interpolation (this induces negligible blur which does not affect the visual quality).

The local geometries preferred by different compression schemes are of different nature. In some cases, those preferred structures look quite unnatural to the human eye (\eg blockiness effects in JPEG). To prevent our approach from generating unpleasant images, we tuned the parameter $\alpha$ of the regularization map \eqref{eq:RegularizationMap} differently for different compression schemes. For JPEG we used $\alpha=20$, for JPEG~2000 and Global and Subband thresholding we used $\alpha=3$, and for WebP, BPG and Deep Coding we used $\alpha=6$. For all compression methods we used $\lambda = 65$ (for color images with 8 bits per pixel per channel). 

The running time of our algorithm is given by \mbox{$ T = K\times(T_{\text{CODEC}}+T_{\text{flow}})$}, where $K$ is the number of iterations and $T_{\text{CODEC}},T_{\text{flow}}$ are the running times of the CODEC and optical flow, respectively. Typically, $K$ is on the order of a few tens, $T_{\text{flow}}\approx0.5$~sec and $T_{\text{CODEC}}\approx0.15$~sec for a 1 megapixel image.

Figures \ref{Fig:DragonFly} and \ref{Fig:Pattern} depict results produced by our algorithm with the Subband Thresholding \cite{strang1996wavelets} and the Global Thresholding \cite{misiti2013wavelets} compression methods, respectively. These two simple approaches produce unpleasing visual results already at moderate compression ratios. However, by using our approach, we are able to improve their performance. This demonstrates that the choice of the error criterion is 
not less important than the choice of the image prior. Indeed, even simple 
models can lead to good visual results 
when used with a deformation indifferent error criterion.

Figures \ref{Fig:Figure1}, \ref{Fig:JPEG2000} and \ref{Fig:JPEG} show several results produced by our algorithm with the JPEG and JPEG~2000 schemes.
As can be seen, our algorithm has a very pronounced visual effect: It manages to preserve a lot of the content that is completely lost in regular compression. In particular see the house's roof and windows and the threads on the leaf.

Next, we applied our algorithm on the newer compression methods WebP and BPG. Generally, we found that the improvement for those methods is moderate and sometimes even unnoticeable. However, in some cases (\eg Figs.~\ref{Fig:WebP} and~\ref{Fig:BPG}) the contribution of our approach is extremely meaningful. Note how our algorithm restores fine details like the words `NEW YORK' and the texture on the sweater, that are otherwise completely dissolved. 

Recently, several neural net based lossy compression methods have been proposed \cite{waveone2017,toderici2016full,johnston2017improved,balle2016end,theis2017lossy,toderici2015variable}. To test the effect of deformation awareness on this family of techniques, we experimented with the CODEC of
\cite{toderici2016full}. As illustrated in Fig.~\ref{Fig:NN}, our approach significantly boosts the visual quality of this method. This suggests that our method is also of great relevance to the recent trend of deep-net based compression.

\begin{figure}[]
	\centering 	
	\includegraphics[width=1\columnwidth, trim=0cm 0cm 0cm 0cm, clip]{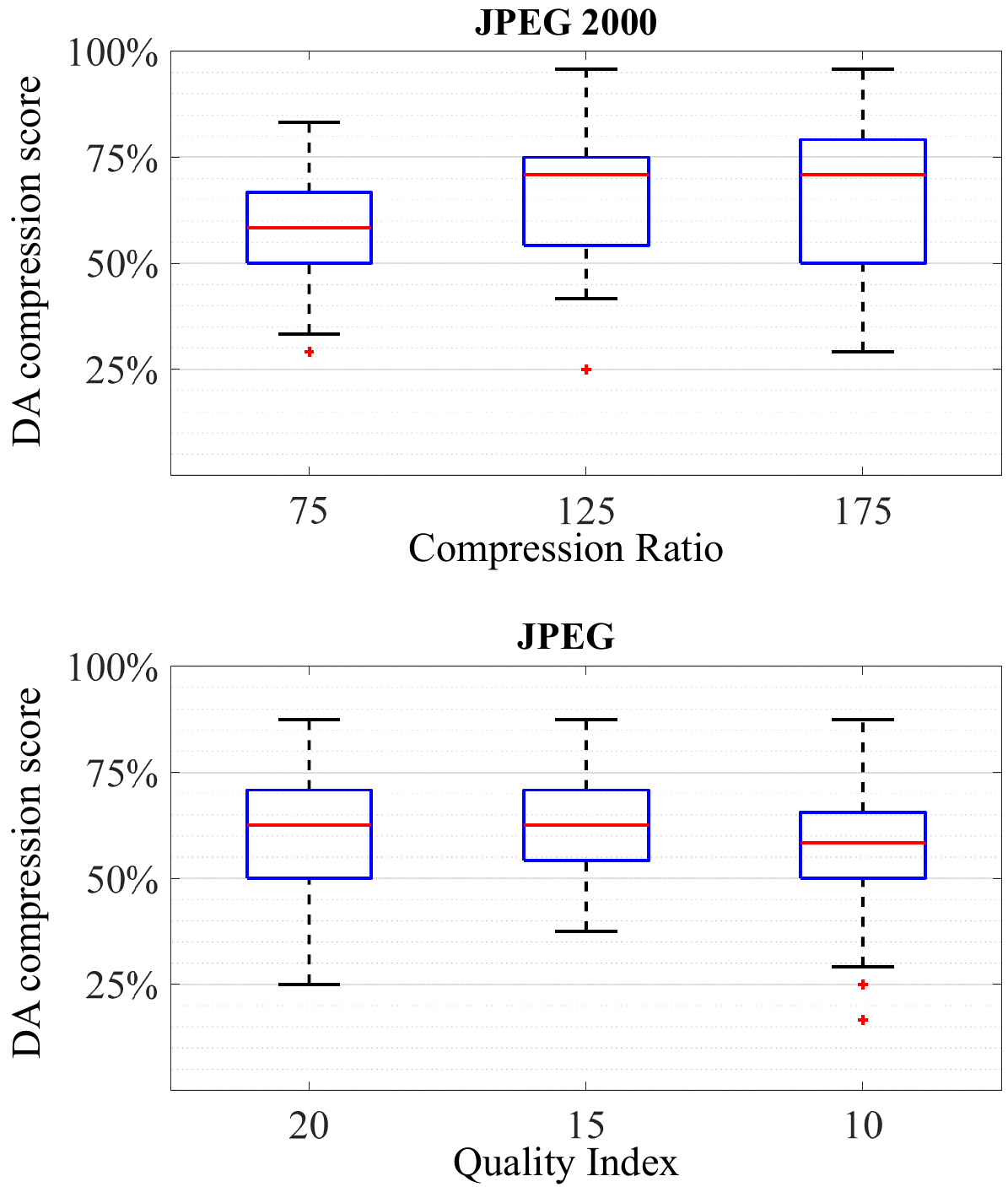}
	\caption{{\textbf{User Study.}
			For each human subject, we recorded the percentage of times he/she preferred the image that was compressed with our deformation aware (DA) versions of JPEG 2000 and JPEG, over that which was compressed with the original methods (see text for details). For each compression ratio in JPEG 2000 and quality factor in JPEG, we plot the median percentage of preference (red line), the 25\% and 75\% percentiles (blue box), the extreme values (black lines), and outliers according to the interquartile ranges (IQR) (red marks). As can be seen, well above 75\% of the subjects preferred our deformation aware version for more than 50\% of the images.
	}}
	\label{fig:UserStudy}
\end{figure}

As demonstrated in Fig.~\ref{Fig:DAMSE}, since our method introduces deformations, most error criteria 
tend to rank our results as less similar to the input image than the regular compression. Therefore, to quantify the perceptual effect of our approach, we conducted a user study on the Kodak dataset~\cite{Kodak}. For each of the 24 uncompressed images in this dataset, the participants were asked to choose which of its two compressed versions looks better: the one with regular compression (with JPEG or JPEG~2000) or our deformation aware variant of the same compression method. For JPEG~2000, the two compared images were compressed with the same ratio (we tested ratios of 75, 125 and 175) and for JPEG both images were compressed with the same quality factor (we tested qualities of 20, 15 and 10). In the case of JPEG, our method resulted in a minor decrease in the compression ratio ($\sim 5\%$ on average). The JPEG~2000 and JPEG surveys were completed by 57 and 59 Amazon Mechanical Turk workers, respectively. 
As can be seen in Fig.~\ref{fig:UserStudy}, the vast majority of the subjects chose our compressed images well above 50\% of times. According to the Wilson test with confidence level of 95\%, the mean percentage of preference for our method is $58.3\% \in [51.7\%,79.9\%]$, $65.9\% \in [60.4\%, 87.6\%]$, $66\% \in [60.5\%,87.7\%]$ for JPEG2000 with compression of ratios of 75, 125, 175, respectively, and $59.7\% \in [53.3\%,80.8\%]$, $62.4\%\in[56.3\%,83.5\%]$, $56.9\% \in [50.2\%,78.0\%]$ for JPEG with quality indices of 20, 15, 25, respectively. This supports the conclusion that the average percentage of preference to our method is well above 50\%, with very high confidence. This indicates that our deformation aware framework indeed leads to a meaningful improvement in visual quality over the original compression methods.


%
\section{Conclusions}
We proposed a generic approach for improving the visual quality of lossy image compression schemes. Our method relies on a new error criterion, which is insensitive to 
smooth deformations. The advantages of our criterion are twofold. First, as opposed to other criteria, it can be easily incorporated into any existing compression scheme. Second, to excel under our criterion, the encoder need not invest bits in describing the exact geometries of fine 
structures. The effect is that more bits are invested in the important parts, leading to 
greatly better detail preservation. User studies confirmed that our approach significantly improves the visual quality of existing compression techniques. 



\noindent\textbf{Acknowledgements\ \ \ }
This research was supported in part by an Alon Fellowship, by the Israel Science Foundation (grant no.~852/17), and by the Ollendorf Foundation.

{	\small
	\bibliographystyle{ieee}
	\bibliography{DeformationAwareCompression}
}

\end{document}